\pdfoutput=1

\documentclass[11pt]{article}

\usepackage[]{acl}

\usepackage{times}
\usepackage{latexsym}

\usepackage[T1]{fontenc}

\usepackage[utf8]{inputenc}

\usepackage{microtype}

\usepackage{inconsolata}

\usepackage{graphicx}

\usepackage{hyperref}

\title{Facts Fade Fast: Evaluating Memorization of Outdated Medical Knowledge in Large Language Models}

\author{Juraj Vladika, Mahdi Dhaini,  Florian Matthes \\
  Technical University of Munich, Germany \\
  School of Computation, Information and Technology \\
  Department of Computer Science \\
  \texttt{\{juraj.vladika, mahdi.dhaini, matthes\}@tum.de}}

\begin{document}
\maketitle
\begin{abstract}
The growing capabilities of Large Language Models (LLMs) show significant potential to enhance healthcare by assisting medical researchers and physicians. However, their reliance on static training data is a major risk when medical recommendations evolve with new research and developments. When LLMs memorize outdated medical knowledge, they can provide harmful advice or fail at clinical reasoning tasks. 
To investigate this problem, we introduce two novel question-answering (QA) datasets derived from systematic reviews: \textit{MedRevQA} (16,501 QA pairs covering general biomedical knowledge) and \textit{MedChangeQA} (a subset of 512 QA pairs where medical consensus has changed over time). 
Our evaluation of eight prominent LLMs on the datasets reveals consistent reliance on outdated knowledge across all models. 
We additionally analyze the influence of obsolete pre-training data and training strategies to explain this phenomenon and propose future directions for mitigation, laying the groundwork for developing more current and reliable medical AI systems.
\end{abstract}

\section{Introduction}

The advent of pre-trained Large Language Models (LLMs) has revolutionized the field of Natural Language Processing (NLP) \cite{naveed2025comprehensive}. One of their most promising application domains is healthcare, where they hold the potential to democratize access to health services and assist in crucial clinical workflows \cite{thirunavukarasu2023large, 10.1001/jamainternmed.2023.1838, liu2025application}.


LLMs are trained to predict the next token on massive amounts of text data, which results in deeply encoding a lot of knowledge in their weights \cite{dhingra2022time, chang2024large}. 
Recent studies suggest that LLMs encode clinical knowledge effectively \cite{singhal2023large, zhang2025much}, by being trained on medical texts like patient records and clinical guidelines. The model's ability to recall specific facts from this data is often referred to as \textit{memorization} \cite{carlini2022quantifying}.

\begin{figure}
    \centering
    \includegraphics[width=0.99\linewidth]{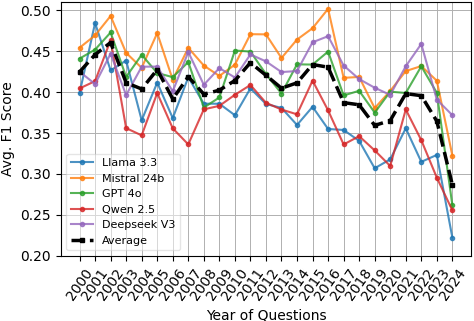}
    \caption{Average F1 scores of five LLMs on medical questions originating from different years in our dataset. The performance decline as questions get more recent points to stronger memorization of older knowledge.}
    \label{fig:f1_years}
\end{figure}


World knowledge quickly evolves in dynamic domains like entertainment or politics. However, this also happens with scientific knowledge. In medicine, new high-quality evidence constantly emerges, rendering previous recommendations obsolete \cite{hodder2024living}. Consequently, the knowledge memorized by an LLM at its training time can become outdated, as they struggle to keep up with the evolving world knowledge \cite{zhang-etal-2023-large}. This is a major safety concern, as it can lead to LLMs providing consumers with incorrect health advice \cite{li2023chatdoctor, ong2024nejm} or fail in clinical settings when using flawed facts in their reasoning \cite{hager2024evaluation}. Even when augmented with retrieved up-to-date information, LLMs can reject it and resort to internal knowledge in so-called \textit{knowledge conflicts} \cite{xu-etal-2024-knowledge-conflicts}.


While recent work has explored the memorization of outdated knowledge in the encyclopedic domain \cite{vu-etal-2024-freshllms, chenghaozhu-etal-2025-llm}, the temporal decay of \textit{medical} knowledge has been less explored. To address this critical gap, we construct new datasets and use them to evaluate memorization of outdated medical knowledge:

\begin{itemize}
    \item We introduce \textbf{MedRevQA}, a new dataset of 16,501 QA pairs from medical systematic reviews; and \textbf{MedChangeQA}, a subset of 512 pairs where answers have changed over time.
    \item We benchmark eight LLMs on our datasets, demonstrating that all models exhibit memorization of outdated medical information.
    \item We provide in-depth analysis, including tracing outdated knowledge to the training data, and discuss promising mitigation strategies.
\end{itemize}


The datasets and code are publicly available.\footnote{\url{https://github.com/jvladika/MedChange}}

\section{Related Work}
NLP has a wide array of applications to the biomedical field \cite{wang2023pre} and LLMs have shown great potential in various medical tasks and clinical applications \cite{thirunavukarasu2023large}. 
A popular task within NLP for healthcare is biomedical question answering (BQA) \cite{jin2022biomedical, nentidis2024overview}.
BQA is seen as a good proxy for evaluating how well LLMs encode and recall medical knowledge \cite{subramanian-etal-2024-qalm, singhal2023large, singhal2025toward} -- therefore, we use it as our main task. The most similar QA dataset in construction is MedREQAL \cite{vladika-etal-2024-medreqal}, but we majorly expand the scope and the purpose.


Recent work has explored how to measure memorized training examples in LLMs \cite{jagielski2023measuring, maini2024llm, kassem-etal-2025-alpaca}. Similarly, temporal QA datasets have been constructed to investigate quickly changing knowledge, mostly focusing on the general, encyclopedic domain \cite{kasai2023realtime, vu-etal-2024-freshllms, 10.1609/aaai.v38i17.29822}.


To the best of our knowledge, we introduce the first QA dataset focusing on knowledge change specifically for the \textit{medical} domain and the first investigation of how much outdated medical knowledge popular LLMs encode.

\section{Dataset}
\paragraph{Systematic Literature Reviews (SLRs).}
Our dateset is constructed from medical SLRs, studies which aim to bring evidence together to answer a pre-defined research question. This involves the identification of primary research relevant to the question, the critical appraisal of this research, and the synthesis of the findings \cite{kolaski2023guidance}.
SLRs are considered the highest quality evidence in the medical "hierarchy of evidence" \cite{wallace2022hierarchy}. We use SLRs to construct a QA dataset because their clear structure and strict criteria used for decisions make them a well-suited proxy for evaluating the state of encoded medical knowledge. We use the SLRs from Cochrane Collaboration \cite{cumpston2022strengthening}, which is the most well-known international organization specializing in the construction of SLRs. Many Cochrane SLRs are updated as new evidence for a question arises.


\begin{table}[t]
\centering

\begin{tabular}{p{\columnwidth}}
\\
\hline
\textbf{Question:} \small Does long-term antibiotic use help prevent recurrent urinary tract infections in children?
 \\ 

\textbf{Conclusion:} \small Long-term antibiotics may reduce the risk of repeat symptomatic UTI in children who have had one or more previous UTIs but the benefit may be small and must be considered together with the increased risk of microbial resistance. (...) [Williams, 2019]
 \\

\textbf{Verdict:}  \textcolor{teal}{\textbf{Supported}} \\
\hline
\textbf{Question:} \small Does long-term antibiotic use help prevent recurrent urinary tract infections in children? \\

\textbf{Conclusion:} \small Large, properly randomised, double blinded studies are needed to determine the efficacy of long-term antibiotics for the prevention of UTI in susceptible children. (...) [Williams, 2011]
 \\

\textbf{Verdict:}  \textcolor{blue}{\textbf{Not Enough Information}} \\
\hline
\end{tabular}

\caption{\label{tab:claims}Example of two instances from our dataset, showing how the verdict changed through time as new, higher quality evidence was discovered.}
\end{table}

\paragraph{Dataset Construction.}
PubMed, the largest database of medical publications \cite{white2020pubmed}, contains all the Cochrane systematic review abstracts from 2000 to 2024 (until January 2024, when we scraped). 
We built a Python scraping script using BeautifulSoup and scraped all the Cochrane SLR abstracts.\footnote{\url{https://pubmed.ncbi.nlm.nih.gov/?term=\%22Cochrane+Database+syst+rev\%22\%5BJournal\%5D}}
Every SLR in the dataset consists of the same sections: \textit{Background, Objectives, Search methods, Selection criteria, Data collection and analysis, Main results,} and \textit{Authors' conclusions}. Our final QA dataset consists of \textbf{questions} and \textbf{labels}. We used \textit{gpt-4o-mini-2024-07-18} to semi-automatically construct the questions and labels, by providing it with the full SLR abstract. 

Questions were derived from the \textit{Objectives} section, by rewriting them to the interrogative form. Labels originate from the \textit{Authors' conclusions} section --
the LLM selects one of the labels \textcolor{teal}{Supported}, \textcolor{red}{Refuted}, or \textcolor{blue}{Not Enough Information} (NEI), as the final \textbf{label}. These labels align with common labels in other medical QA and fact-checking datasets \cite{glockner-etal-2024-ambifc, vladika-etal-2024-healthfc}.
In total, this dataset has 16,501 QA pairs, spanning virtually all medical disciplines and covering a wide array of important biomedical questions for benchmarking. We call this dataset \textbf{MedRevQA}.

\paragraph{Changed Knowledge.}
Our dataset consists of 16,501 SLR records. Out of those, 12,122 are unique SLRs that have never had an update. The remaining 4379 SLRs constitute 1535 groups (with a minimum of 2 SLRs in a group, a maximum of 9, and a mean of 2.85) that researched the same question. This means there are 1535 research questions that have had multiple SLR iterations written about them. Out of 1535, \textbf{512} have had a verdict change over time, meaning that the authors changed the conclusion of the investigated research question in a follow-up SLR study, when they acquired updated evidence from research. This follows findings from medical research studies that have shown how 20 to 30\% of Cochrane reviews change their conclusions throughout time \cite{hughes2012cochrane, https://doi.org/10.1002/jrsm.1556}.
We consolidate these questions with changed verdicts into the \textbf{MedChangeQA} dataset and collect all their verdicts through different iterations. MedChangeQA has \textbf{questions}, \textbf{latest label}, and (the most recent) \textbf{outdated label} for those studies where the label changed.

\paragraph{Annotation Quality.} Two annotators, one who is our in-house physician from the university clinic and another an author with a background in biomedical engineering, evaluated a random subset of 100 examples for the correctness of generated questions and verdicts. They found 95\% of questions and 92\% of labels to be correct. We deem this to be relatively high, since even the human annotation process is imperfect \cite{klie2023annotation}, with errors due to incorrect problem understanding or loss of concentration. A common source of label errors was conflating Refuted and NEI labels. On the other hand, all 512 labels in MedChangeQA were manually checked and corrected by the two annotators. Therefore, MedRevQA has silver labels, while MedChangeQA has gold labels.

\paragraph{Dataset Description.}
In total, MedRevQA has 16,501 questions, of which 6499 are \textit{Supported}, 3124 are \textit{Refuted}, and for 6878 there is \textit{Not enough information}. In MedChangeQA, for the 512 questions with changed verdicts, the newest labels have a 221/131/160 ratio for S/R/NEI, and the outdated labels are at 152/123/237 for S/R/NEI, showing how the most common change is from not having enough information to becoming supported or refuted by relevant research. Still, some questions can go from support/refute to inconclusive findings with more research (see an example in Table \ref{tab:citewrong}).

\begin{table*}[htpb]
\centering
\small
\resizebox{\linewidth}{!}{%
\begin{tabular}{l|c|ccc||ccc|ccc|c|c}

\hline
\multicolumn{2}{c}{ } & \multicolumn{3}{c}{(a) \textbf{Full Dataset}~(16.5k)} & & \multicolumn{6}{c}{\textbf{Changed Knowledge Dataset} (512)} &  \\ \hline
& \textbf{Release} & & & & \multicolumn{3}{c}{(b)~\textbf{Outdated Lab.}~$\downarrow$}  & \multicolumn{3}{c}{(c)~\textbf{Latest Labels}~$\uparrow$} & \textbf{F1} & \textbf{Outdated}\\
     & \textbf{Date} & \textbf{P} & \textbf{R}  & \textbf{F1} & \textbf{P} & \textbf{R}  & \textbf{F1}  & \textbf{P} & \textbf{R}  & \textbf{F1} & \textbf{diff.} & \textbf{Answers}  \\ \hline
\textbf{GPT-4o}  &  2024-05-13  & 52.6 & 45.1 & 42.9 & 45.5 & 38.9 & 34.1 & 35.2 & 34.5 & 31.1 & \textcolor{red}{-3.0}  & 39.4\% \\ 
\textbf{Mistral 24B}  & 2025-01-30 & 50.6 & \textbf{46.3} & \textbf{45.7} & 38.2 & 37.6 & 33.9 & 36.9 & 35.5 & 33.7 & \textcolor{red}{-0.2} & 38.7\% \\ 
\textbf{Llama 3.3 70B}  &  2024-12-06  & 52.7 & 45.9 & 39.3 & 38.9 & 36.6 & \textbf{26.7} & \textbf{42.8} & \textbf{39.3} & 34.1 & \textbf{\textcolor{teal}{+7.4}} & 32.2\% \\ 
\textbf{Qwen 2.5 7B}  &  2024-09-19  & 46.4 & 42.3 & 38.7 & 42.6 & 37.1 & 30.8 & 27.1 & 30.8 & 26.0 & \textbf{\textcolor{red}{-4.8}}   & 35.4\% \\ 
\textbf{Deepseek V3} & 2024-12-26 & \textbf{56.2} & 46.2 & 43.8 & 43.2 & 38.6 & 33.9 & 40.2 & 35.1 & 32.2 & \textcolor{red}{-1.7}   & 40.6\%  \\ 
\textbf{OLMo 2 13B} & 2024-11-24 & 43.5 & 42.5 & 37.9 & \textbf{36.2} & \textbf{35.3} & 29.3 & 35.5 & 35.7 & 33.2 & \textcolor{teal}{+2.9}   & 32.0\% \\ \hline
\textbf{PMC-Lm 13B} & 2023-08-28  & 39.5 & 37.6 & 36.5 & 41.9 & 39.8 & 35.9 & 34.5 & 34.3 & 33.1 & \textcolor{red}{-2.8}   & 37.3\% \\ 
\textbf{BioMistral 7B} & 2024-02-19 & 41.2 & 41.5 & 40.9 & 36.8 & 37.2 & 36.3 & 35.4 & 35.5 & \textbf{35.3}  & \textcolor{red}{-1.5}   & 37.1\%  \\ \hline
\end{tabular}
}

\caption{\label{tab:performance-results} Final results of eight LLMs, measured by macro Precision (P), Recall (R), and F1. Experiments include (a) the full dataset; and the changed knowledge dataset, using (b) outdated labels and (c) latest labels as ground truth, respectively. 
The final column is the percentage of answers in (b) where an outdated label was predicted.
}
\end{table*}

\section{Experimental Setup}
The experiments consisted of instructing the LLMs to predict one of the three labels (S/R/NEI), given the medical question as input. No additional context was provided, as the goal is to evaluate their internal knowledge and memorization. The models also explained their output.
For evaluation, we extract the predicted label and compare it to ground-truth labels from the dataset, using the (macro-averaged) precision, recall, and F1 scores.

We test multiple LLMs, starting with \textbf{GPT-4o} (\textit{2024-08-06}), as the most popular proprietary LLM. We also benchmark four open-weights LLMs: \textbf{Mistral 24B}, \textbf{Llama 3.3 (70B)}, \textbf{Qwen 2.5 (7B)}, \textbf{DeepSeek-V3} (\textbf{685B}); and finally the fully open-source \textbf{OLMo 2 (13B)}. See Appendix \ref{sec:training_data} for a summary of public info on their pre-training data.

To compare general-purpose LLMs to domain-specific ones, we also benchmark the performance of \textbf{PMC-LLaMa 13B} \cite{wu2023pmcllama}, an extension of Llama 2, 
and \textbf{BioMistral 7B} \cite{labrak-etal-2024-biomistral}, an extension of Mistral-v0.2; both further pre-trained on biomedical research papers.


All prompts can be found in Table~\ref{tab:llm_prompts}. GPT 4o was prompted through the OpenAI API. The four general-purpose models were prompted via the API of Together AI. Two biomedical LLMs were run locally (in an 8-bit quantized version) on one Nvidia V100 GPU with 16 GB VRAM, for two computation hours. The token limit was set to 512 and the temperature to 0 to maximize deterministic outputs.



\section{Results}
\noindent \paragraph{Experiment Rounds.} We first test (a) the full dataset, MedRevQA.
We also did two experiments on MedChangeQA,
first with (b) outdated labels as ground truth, then with (c) latest labels as ground truth. 
We use the difference between the scores of (b) and (c) as a proxy to show the extent of outdated medical knowledge in LLMs.
Final results are systematized in Table~\ref{tab:performance-results}, measured by macro P, R, and F1.
The last column shows the percentage of answers in the 3rd experiment (c) where the outdated label was predicted (and not the correct latest label or an incorrect label altogether). 

\noindent \paragraph{Performance.} On the full dataset, Mistral exhibited the best R and F1, showing it has the best overview of the overall medical knowledge landscape. Precision was the highest in Deepseek-V3. Nevertheless, none of the models has a very high performance, pointing to the challenging nature of MedRevQA as a general biomedical QA testbed.

When it comes to outdated knowledge, Llama 3.3 had the highest degree of the latest knowledge as compared to the outdated labels (+7.4), while OLMo also had a positive difference (+2.9). Mistral showed an almost identical performance, while GPT, Qwen, DeepSeek, and PMC-Llama all struggled. Qwen was also the smallest and least capable model, which could explain low scores in general and low awareness of recent knowledge.

An example of outdated and incorrect knowledge is shown in Table~\ref{tab:outdated} in Appendix. Additionally, Figure \ref{fig:f1_years}/\ref{fig:full_f1_years} shows how the average F1 across LLMs on questions from different years on MedRevQA declines in more recent years, as all post-2016 average scores are lower than any beforehand. A similar drop in LLM performance on more recent medical questions was found by \citet{park2025chroknowledge}.

\section{Discussion and Analysis}

\begin{figure}[htpb]
    \centering
    \includegraphics[width=0.9\linewidth]{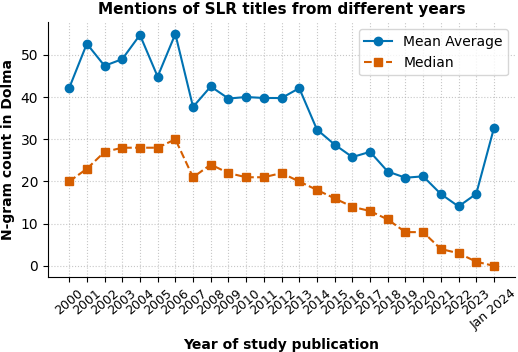}
    \caption{N-gram counts per the year of studies in Dolma, the pre-training corpus of OLMo.}
    \label{fig:ngrams}
\end{figure}

\noindent \paragraph{Pre-training Data.} Most popular LLMs do not fully disclose their pre-training data, making it difficult to assert if concrete medical studies were memorized. 
Still, recent studies demonstrated empirically the presence of memorized medical datasets \cite{gallifant-etal-2024-language, yang2024memorization}.
We also saw a tendency of models to explicitly mention specific studies in their answers, including Cochrane reviews, many of which were decade-old (see Table \ref{tab:citewrong}), thus displaying outdated memorized knowledge (see Table \ref{tab:mentions}). 
We outline pre-training corpora of used LLMs in Appendix \ref{sec:training_data}, and for the fully open OLMo, we show the presence of all used SLRs in its pre-training corpus, with earlier ones being more prevalent (Figure \ref{fig:ngrams}). 

\paragraph{Inspection of OLMo.}
OLMo 2 \cite{olmo20252olmo2furious} is trained on the Dolma corpus \cite{soldaini-etal-2024-dolma}, a fully open dataset containing around 3 trillion tokens. 
It contains the peS2o \cite{peS2o} and S2ORC \cite{lo-etal-2020-s2orc} corpora that constitute the academic knowledge base Semantic Scholar. This database also indexes all of Cochrane's systematic literature reviews.\footnote{\url{https://www.semanticscholar.org/venue?name=Cochrane\%20Database\%20of\%20Systematic\%20Reviews}}
Therefore, we can with high certainty say that the OLMo models have seen Cochrane's SLRs during its pre-training. Other than in the two academic corpora, there is a wide presence of these SLRs in other parts of the dataset, especially various online websites found in Common Crawl \cite{dodge-etal-2021-documenting}. 

We used Infini-gram \cite{liu2024infinigram}, an n-gram language model that can be used to query Dolma and other pre-training corpora,\footnote{\url{https://huggingface.co/spaces/liujch1998/infini-gram}} to inspect the presence of Cochrane's SLRs. Searching for "\textit{Cochrane Database of Systematic Reviews}" (the exact journal name, case-sensitive) returns 144,493 hits for Dolma v1.7 (used for OLMo 2). Additionally, we queried the title of each of the SLR studies found in MedRevQA and report on the mean and median amount of n-gram counts per year in Figure \ref{fig:ngrams}. The mean and median almost steadily decrease over years, meaning that the most mentioned and discussed studies are the earliest ones since they have had more time to spread throughout the web. The higher frequency of mentions can lead to to stronger encoding of outdated knowledge in LLMs.

\paragraph{Mentions of Specific Studies}
In Table \ref{tab:mentions}, we show the number of mentions of some common terms referring to specific medical studies (such as \textit{systematic review}, \textit{meta-analysis}, \textit{journal}), across all LLM answers on MedRevQA questions. This shows how models tend to cite specific studies when providing some of their answers, which is useful for source attribution, but becomes problematic when the referred studies are outdated and deprecated. 
It is notable how GPT overwhelmingly resorted to using general phrasing such as "\textit{studies have shown a positive effect...}" without specifying what studies exactly it is referring to. This likely comes from its final alignment and preference-learning phase, where a particular answering style is learned.

\noindent \paragraph{Potential Explanations.} We hypothesize some reasons for the presence of strongly encoded outdated knowledge. 
Firstly, older scientific findings have been around for a longer time and have already permeated the Internet, news, follow-up studies, and are present in pre-training corpora.
Additionally, scientific findings are often misrepresented online \cite{glockner2024missci, wuhrl2024understanding}, so faulty medical knowledge could get encoded.

\begin{figure}[htpb]
    \centering
    \includegraphics[width=0.95\linewidth]{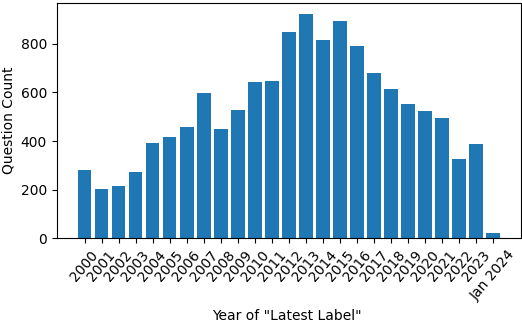}
    \caption{Distribution of the year of "Latest Label" -- the label of the most recent study for a given question}
    \label{fig:label_years}
\end{figure}

Secondly, LLM memorization rate has been correlated in past work with various training parameters, such as learning rate \cite{tirumala2022memorization}, model size \cite{biderman2023emergent}, or frequency of appearance in training data \cite{carlini2023quantifying}. 
Therefore, it is possible that Llama had the highest data quality and more weight during training put on more recent text, leading to less outdated knowledge. 
Finally, the cutoff of all models is 2023, and the vast majority of "latest labels" are from before 2023 (see Figure~\ref{fig:label_years}). Cutoff could explain the drop in 2023/2024 (Figure~\ref{fig:f1_years}) but not earlier years.

\noindent \paragraph{Future Directions.} One way to overcome outdated knowledge is using retrieval-augmented generation (RAG). We show in Appendix \ref{sec:appendix_rag} how a simple retrieval strategy (with the first PubMed result added to prompt) can already bring improvements. Still, LLMs can hallucinate extra information even in RAG settings \cite{adlakha2024tacl} or not follow the provided references \cite{liu2023evaluating}. Therefore, advanced RAG techniques are needed \cite{yu2024rankrag}, including filtering and re-ranking of retrieved evidence by recency and source quality. 
Future work could also investigate more nuanced labels or long-form answer generation with explanations.
Other promising directions for mitigation of memorized knowledge include: resolving knowledge conflicts \cite{wang2024resolving}, machine \textit{unlearning} \cite{NEURIPS2024_be52acf6, gao2025on}, knowledge editing \cite{wang2024knowledge, jiang2024learning}, and continual learning \cite{shi2024continual}. 



\section{Conclusion}
We introduce two new QA datasets constructed from rigorous biomedical SLRs for benchmarking the general biomedical knowledge of LLMs. With a subset of 512 questions where the answer (verdict) changed over the years, 
we showcase how eight popular LLMs fare better on older medical knowledge and encode a considerable amount of outdated knowledge labels, which can hinder their usability in healthcare settings, including helping physicians, researchers, and patients.  
We outline future work directions and hope our datasets will serve as a challenging testbed for tackling the LLM memorization of outdated knowledge.  



\section*{Limitations}
Most of the MedRevQA dataset introduced in this study was constructed semi-automatically, by scraping the content and using an LLM to generate the question and label. It is possible that some of the generated questions and labels are imperfect. Our manual analysis of 100 randomly selected instances showed that the performance is 92--95\% correct, for labels and questions, respectively. We considered this to be a good enough performance, considering that even human annotation is not always perfect. MedRevQA should be interpreted as having silver labels and used as such.

We use the difference in F1 scores between the predicted labels when using "outdated labels" and "latest labels" as ground truth, as a proxy for evaluating the degree of encoded outdated medical knowledge. This is not a perfect measure since it is possible that the LLM predicted an incorrect label due to some logical error or misinterpreting the question. Still, our manual inspection of a large number of generated labels and explanations showed that outdated references were indeed the most common explanation for the label misprediction, and models often referred to old SLRs and meta-analyses, dating back many years.


Given the quickly evolving landscape of new LLMs, some of the LLMs we evaluate can get outdated and deprecated quickly. Additionally, we do not test all the relevant models since some were not available or computationally too expensive for us to run. Due to a lack of resources, our study also lacks a deeper human evaluation of the generated model labels with medical experts, which could have given a more rigorous evaluation.

\section*{Ethics Statement}
The work presented in this study focuses on the sensitive fields of healthcare and medical AI. We predict answers to questions in a zero-shot setting to uncover their internal encoded medical knowledge for research purposes, but this is not suitable for end users or patients. Some responses can include inaccuracies and misleading medical advice, which should be critically evaluated and verified with reliable sources or medical professionals.

The original text of Cochrane's systematic literature review abstracts belongs to the Cochrane Collaboration. We will release only the generated questions and label pairs, and respect the copyright of the original text of Cochrane authors.

\bibliography{custom}

\begin{thebibliography}{62}
\providecommand{\natexlab}[1]{#1}

\bibitem[{Adlakha et~al.(2024)Adlakha, BehnamGhader, Lu, Meade, and Reddy}]{adlakha2024tacl}
Vaibhav Adlakha, Parishad BehnamGhader, Xing~Han Lu, Nicholas Meade, and Siva Reddy. 2024.
\newblock \href {https://doi.org/10.1162/tacl_a_00667} {Evaluating correctness and faithfulness of instruction-following models for question answering}.
\newblock \emph{Transactions of the Association for Computational Linguistics}, 12:681--699.

\bibitem[{Ayers et~al.(2023)Ayers, Poliak, Dredze, Leas, Zhu, Kelley, Faix, Goodman, Longhurst, Hogarth, and Smith}]{10.1001/jamainternmed.2023.1838}
John~W. Ayers, Adam Poliak, Mark Dredze, Eric~C. Leas, Zechariah Zhu, Jessica~B. Kelley, Dennis~J. Faix, Aaron~M. Goodman, Christopher~A. Longhurst, Michael Hogarth, and Davey~M. Smith. 2023.
\newblock Comparing physician and artificial intelligence chatbot responses to patient questions posted to a public social media forum.
\newblock \emph{JAMA Internal Medicine}, 183(6):589--596.

\bibitem[{Babić et~al.(2022)Babić, Poklepović~Peričić, Pieper, and Puljak}]{https://doi.org/10.1002/jrsm.1556}
Andrija Babić, Tina Poklepović~Peričić, Dawid Pieper, and Livia Puljak. 2022.
\newblock \href {https://doi.org/10.1002/jrsm.1556} {When is the evidence conclusive? analysis of systematic reviews for which cochrane declared that conclusions will not change with further studies}.
\newblock \emph{Research Synthesis Methods}, 13(4):478--488.

\bibitem[{Biderman et~al.(2023)Biderman, Prashanth, Sutawika, Schoelkopf, Anthony, Purohit, and Raff}]{biderman2023emergent}
Stella Biderman, USVSN~Sai Prashanth, Lintang Sutawika, Hailey Schoelkopf, Quentin~Gregory Anthony, Shivanshu Purohit, and Edward Raff. 2023.
\newblock \href {https://openreview.net/forum?id=Iq0DvhB4Kf} {Emergent and predictable memorization in large language models}.
\newblock In \emph{Thirty-seventh Conference on Neural Information Processing Systems}.

\bibitem[{Carlini et~al.(2022)Carlini, Ippolito, Jagielski, Lee, Tramer, and Zhang}]{carlini2022quantifying}
Nicholas Carlini, Daphne Ippolito, Matthew Jagielski, Katherine Lee, Florian Tramer, and Chiyuan Zhang. 2022.
\newblock Quantifying memorization across neural language models.
\newblock In \emph{The Eleventh International Conference on Learning Representations}.

\bibitem[{Carlini et~al.(2023)Carlini, Ippolito, Jagielski, Lee, Tramer, and Zhang}]{carlini2023quantifying}
Nicholas Carlini, Daphne Ippolito, Matthew Jagielski, Katherine Lee, Florian Tramer, and Chiyuan Zhang. 2023.
\newblock \href {https://openreview.net/forum?id=TatRHT_1cK} {Quantifying memorization across neural language models}.
\newblock In \emph{The Eleventh International Conference on Learning Representations}.

\bibitem[{Chang et~al.(2024)Chang, Park, Ye, Yang, Seo, Chang, and Seo}]{chang2024large}
Hoyeon Chang, Jinho Park, Seonghyeon Ye, Sohee Yang, Youngkyung Seo, Du-Seong Chang, and Minjoon Seo. 2024.
\newblock How do large language models acquire factual knowledge during pretraining?
\newblock \emph{Advances in neural information processing systems}, 37:60626--60668.

\bibitem[{ChenghaoZhu et~al.(2025)ChenghaoZhu, Chen, Gao, Zhang, Tiwari, and Wang}]{chenghaozhu-etal-2025-llm}
ChenghaoZhu ChenghaoZhu, Nuo Chen, Yufei Gao, Yunyi Zhang, Prayag Tiwari, and Benyou Wang. 2025.
\newblock \href {https://aclanthology.org/2025.naacl-long.381/} {Is your {LLM} outdated? a deep look at temporal generalization}.
\newblock In \emph{Proceedings of the 2025 Conference of the Nations of the Americas Chapter of the Association for Computational Linguistics: Human Language Technologies (Volume 1: Long Papers)}, pages 7433--7457, Albuquerque, New Mexico. Association for Computational Linguistics.

\bibitem[{Cumpston et~al.(2022)Cumpston, McKenzie, Welch, and Brennan}]{cumpston2022strengthening}
Miranda~S Cumpston, Joanne~E McKenzie, Vivian~A Welch, and Sue~E Brennan. 2022.
\newblock Strengthening systematic reviews in public health: guidance in the cochrane handbook for systematic reviews of interventions.
\newblock \emph{Journal of Public Health}, 44(4):e588--e592.

\bibitem[{Dhingra et~al.(2022)Dhingra, Cole, Eisenschlos, Gillick, Eisenstein, and Cohen}]{dhingra2022time}
Bhuwan Dhingra, Jeremy~R Cole, Julian~Martin Eisenschlos, Daniel Gillick, Jacob Eisenstein, and William~W Cohen. 2022.
\newblock Time-aware language models as temporal knowledge bases.
\newblock \emph{Transactions of the Association for Computational Linguistics}, 10:257--273.

\bibitem[{Dodge et~al.(2021)Dodge, Sap, Marasovi{\'c}, Agnew, Ilharco, Groeneveld, Mitchell, and Gardner}]{dodge-etal-2021-documenting}
Jesse Dodge, Maarten Sap, Ana Marasovi{\'c}, William Agnew, Gabriel Ilharco, Dirk Groeneveld, Margaret Mitchell, and Matt Gardner. 2021.
\newblock \href {https://doi.org/10.18653/v1/2021.emnlp-main.98} {Documenting large webtext corpora: A case study on the colossal clean crawled corpus}.
\newblock In \emph{Proceedings of the 2021 Conference on Empirical Methods in Natural Language Processing}, pages 1286--1305, Online and Punta Cana, Dominican Republic. Association for Computational Linguistics.

\bibitem[{Gallifant et~al.(2024)Gallifant, Chen, Moreira, Munch, Gao, Pond, Celi, Aerts, Hartvigsen, and Bitterman}]{gallifant-etal-2024-language}
Jack Gallifant, Shan Chen, Pedro Jos{\'e}~Ferreira Moreira, Nikolaj Munch, Mingye Gao, Jackson Pond, Leo~Anthony Celi, Hugo Aerts, Thomas Hartvigsen, and Danielle Bitterman. 2024.
\newblock \href {https://doi.org/10.18653/v1/2024.findings-emnlp.726} {Language models are surprisingly fragile to drug names in biomedical benchmarks}.
\newblock In \emph{Findings of the Association for Computational Linguistics: EMNLP 2024}, pages 12448--12465, Miami, Florida, USA. Association for Computational Linguistics.

\bibitem[{Gao et~al.(2025)Gao, Wang, Ding, Weng, Wang, and Zhu}]{gao2025on}
Chongyang Gao, Lixu Wang, Kaize Ding, Chenkai Weng, Xiao Wang, and Qi~Zhu. 2025.
\newblock \href {https://openreview.net/forum?id=Essg9kb4yx} {On large language model continual unlearning}.
\newblock In \emph{The Thirteenth International Conference on Learning Representations}.

\bibitem[{Glockner et~al.(2024{\natexlab{a}})Glockner, Hou, Nakov, and Gurevych}]{glockner2024missci}
Max Glockner, Yufang Hou, Preslav Nakov, and Iryna Gurevych. 2024{\natexlab{a}}.
\newblock Missci: Reconstructing fallacies in misrepresented science.
\newblock In \emph{Proceedings of the 62nd Annual Meeting of the Association for Computational Linguistics (Volume 1: Long Papers)}, pages 4372--4405.

\bibitem[{Glockner et~al.(2024{\natexlab{b}})Glockner, Stali{\={u}}nait{\.{e}}, Thorne, Vallejo, Vlachos, and Gurevych}]{glockner-etal-2024-ambifc}
Max Glockner, Ieva Stali{\={u}}nait{\.{e}}, James Thorne, Gisela Vallejo, Andreas Vlachos, and Iryna Gurevych. 2024{\natexlab{b}}.
\newblock \href {https://doi.org/10.1162/tacl_a_00629} {{A}mbi{FC}: Fact-checking ambiguous claims with evidence}.
\newblock \emph{Transactions of the Association for Computational Linguistics}, 12:1--18.

\bibitem[{Hager et~al.(2024)Hager, Jungmann, Holland, Bhagat, Hubrecht, Knauer, Vielhauer, Makowski, Braren, Kaissis et~al.}]{hager2024evaluation}
Paul Hager, Friederike Jungmann, Robbie Holland, Kunal Bhagat, Inga Hubrecht, Manuel Knauer, Jakob Vielhauer, Marcus Makowski, Rickmer Braren, Georgios Kaissis, et~al. 2024.
\newblock Evaluation and mitigation of the limitations of large language models in clinical decision-making.
\newblock \emph{Nature medicine}, 30(9):2613--2622.

\bibitem[{Hodder et~al.(2024)Hodder, Vogel, Wolfenden, and Turner}]{hodder2024living}
Rebecca~K Hodder, Joshua~P Vogel, Luke Wolfenden, and Tari Turner. 2024.
\newblock Living systematic reviews and living guidelines to maintain the currency of public health guidelines.
\newblock \emph{American journal of public health}, 114(1):21--26.

\bibitem[{Hughes et~al.(2012)Hughes, van Wely, and Farquhar}]{hughes2012cochrane}
EG~Hughes, M~van Wely, and CM~Farquhar. 2012.
\newblock Cochrane reviews in perspective: the importance of appropriate conclusions and timing of publication.

\bibitem[{Jagielski et~al.(2023)Jagielski, Thakkar, Tramer, Ippolito, Lee, Carlini, Wallace, Song, Thakurta, Papernot, and Zhang}]{jagielski2023measuring}
Matthew Jagielski, Om~Thakkar, Florian Tramer, Daphne Ippolito, Katherine Lee, Nicholas Carlini, Eric Wallace, Shuang Song, Abhradeep~Guha Thakurta, Nicolas Papernot, and Chiyuan Zhang. 2023.
\newblock \href {https://openreview.net/forum?id=7bJizxLKrR} {Measuring forgetting of memorized training examples}.
\newblock In \emph{The Eleventh International Conference on Learning Representations}.

\bibitem[{Jiang et~al.(2024)Jiang, Wang, Wu, Zhong, Zeng, Gao, Li, Jiang, Shang, Tang et~al.}]{jiang2024learning}
Yuxin Jiang, Yufei Wang, Chuhan Wu, Wanjun Zhong, Xingshan Zeng, Jiahui Gao, Liangyou Li, Xin Jiang, Lifeng Shang, Ruiming Tang, et~al. 2024.
\newblock Learning to edit: Aligning llms with knowledge editing.
\newblock In \emph{Proceedings of the 62nd Annual Meeting of the Association for Computational Linguistics (Volume 1: Long Papers)}, pages 4689--4705.

\bibitem[{Jin et~al.(2022)Jin, Yuan, Xiong, Yu, Ying, Tan, Chen, Huang, Liu, and Yu}]{jin2022biomedical}
Qiao Jin, Zheng Yuan, Guangzhi Xiong, Qianlan Yu, Huaiyuan Ying, Chuanqi Tan, Mosha Chen, Songfang Huang, Xiaozhong Liu, and Sheng Yu. 2022.
\newblock Biomedical question answering: a survey of approaches and challenges.
\newblock \emph{ACM Computing Surveys (CSUR)}, 55(2):1--36.

\bibitem[{Kasai et~al.(2023)Kasai, Sakaguchi, yoichi takahashi, Bras, Asai, Yu, Radev, Smith, Choi, and Inui}]{kasai2023realtime}
Jungo Kasai, Keisuke Sakaguchi, yoichi takahashi, Ronan~Le Bras, Akari Asai, Xinyan~Velocity Yu, Dragomir Radev, Noah~A. Smith, Yejin Choi, and Kentaro Inui. 2023.
\newblock \href {https://openreview.net/forum?id=HfKOIPCvsv} {Realtime {QA}: What's the answer right now?}
\newblock In \emph{Thirty-seventh Conference on Neural Information Processing Systems Datasets and Benchmarks Track}.

\bibitem[{Kassem et~al.(2025)Kassem, Mahmoud, Mireshghallah, Kim, Tsvetkov, Choi, Saad, and Rana}]{kassem-etal-2025-alpaca}
Aly~M. Kassem, Omar Mahmoud, Niloofar Mireshghallah, Hyunwoo Kim, Yulia Tsvetkov, Yejin Choi, Sherif Saad, and Santu Rana. 2025.
\newblock \href {https://doi.org/10.18653/v1/2025.naacl-long.421} {{ALPACA} {AGAINST} {VICUNA}: Using {LLM}s to uncover memorization of {LLM}s}.
\newblock In \emph{Proceedings of the 2025 Conference of the Nations of the Americas Chapter of the Association for Computational Linguistics: Human Language Technologies (Volume 1: Long Papers)}, pages 8296--8321, Albuquerque, New Mexico. Association for Computational Linguistics.

\bibitem[{Klie et~al.(2023)Klie, Webber, and Gurevych}]{klie2023annotation}
Jan-Christoph Klie, Bonnie Webber, and Iryna Gurevych. 2023.
\newblock Annotation error detection: Analyzing the past and present for a more coherent future.
\newblock \emph{Computational Linguistics}, 49(1):157--198.

\bibitem[{Kolaski et~al.(2023)Kolaski, Romeiser~Logan, and Ioannidis}]{kolaski2023guidance}
Kat Kolaski, Lynne Romeiser~Logan, and John~PA Ioannidis. 2023.
\newblock Guidance to best tools and practices for systematic reviews.
\newblock \emph{Journal of Pediatric Rehabilitation Medicine}, 16(2):241--273.

\bibitem[{Labrak et~al.(2024)Labrak, Bazoge, Morin, Gourraud, Rouvier, and Dufour}]{labrak-etal-2024-biomistral}
Yanis Labrak, Adrien Bazoge, Emmanuel Morin, Pierre-Antoine Gourraud, Mickael Rouvier, and Richard Dufour. 2024.
\newblock \href {https://doi.org/10.18653/v1/2024.findings-acl.348} {{B}io{M}istral: A collection of open-source pretrained large language models for medical domains}.
\newblock In \emph{Findings of the Association for Computational Linguistics: ACL 2024}, pages 5848--5864, Bangkok, Thailand. Association for Computational Linguistics.

\bibitem[{Li et~al.(2024)Li, Guerin, and Lin}]{10.1609/aaai.v38i17.29822}
Yucheng Li, Frank Guerin, and Chenghua Lin. 2024.
\newblock \href {https://doi.org/10.1609/aaai.v38i17.29822} {Latesteval: addressing data contamination in language model evaluation through dynamic and time-sensitive test construction}.
\newblock In \emph{Proceedings of the Thirty-Eighth AAAI Conference on Artificial Intelligence and Thirty-Sixth Conference on Innovative Applications of Artificial Intelligence and Fourteenth Symposium on Educational Advances in Artificial Intelligence}, AAAI'24/IAAI'24/EAAI'24. AAAI Press.

\bibitem[{Li et~al.(2023)Li, Li, Zhang, Dan, Jiang, and Zhang}]{li2023chatdoctor}
Yunxiang Li, Zihan Li, Kai Zhang, Ruilong Dan, Steve Jiang, and You Zhang. 2023.
\newblock Chatdoctor: A medical chat model fine-tuned on a large language model meta-ai (llama) using medical domain knowledge.
\newblock \emph{Cureus}, 15(6).

\bibitem[{Liu et~al.(2025)Liu, Zhou, Gu, Zou, Huang, Wu, Li, Chen, Hua, Zhou et~al.}]{liu2025application}
Fenglin Liu, Hongjian Zhou, Boyang Gu, Xinyu Zou, Jinfa Huang, Jinge Wu, Yiru Li, Sam~S Chen, Yining Hua, Peilin Zhou, et~al. 2025.
\newblock Application of large language models in medicine.
\newblock \emph{Nature Reviews Bioengineering}, pages 1--20.

\bibitem[{Liu et~al.(2024)Liu, Min, Zettlemoyer, Choi, and Hajishirzi}]{liu2024infinigram}
Jiacheng Liu, Sewon Min, Luke Zettlemoyer, Yejin Choi, and Hannaneh Hajishirzi. 2024.
\newblock \href {https://openreview.net/forum?id=u2vAyMeLMm} {Infini-gram: Scaling unbounded n-gram language models to a trillion tokens}.
\newblock In \emph{First Conference on Language Modeling}.

\bibitem[{Liu et~al.(2023)Liu, Zhang, and Liang}]{liu2023evaluating}
Nelson~F Liu, Tianyi Zhang, and Percy Liang. 2023.
\newblock Evaluating verifiability in generative search engines.
\newblock In \emph{Findings of the Association for Computational Linguistics: EMNLP 2023}, pages 7001--7025.

\bibitem[{Lo et~al.(2020)Lo, Wang, Neumann, Kinney, and Weld}]{lo-etal-2020-s2orc}
Kyle Lo, Lucy~Lu Wang, Mark Neumann, Rodney Kinney, and Daniel Weld. 2020.
\newblock \href {https://doi.org/10.18653/v1/2020.acl-main.447} {{S}2{ORC}: The semantic scholar open research corpus}.
\newblock In \emph{Proceedings of the 58th Annual Meeting of the Association for Computational Linguistics}, pages 4969--4983, Online. Association for Computational Linguistics.

\bibitem[{Maini et~al.(2024)Maini, Jia, Papernot, and Dziedzic}]{maini2024llm}
Pratyush Maini, Hengrui Jia, Nicolas Papernot, and Adam Dziedzic. 2024.
\newblock \href {https://openreview.net/forum?id=Fr9d1UMc37} {{LLM} dataset inference: Did you train on my dataset?}
\newblock In \emph{The Thirty-eighth Annual Conference on Neural Information Processing Systems}.

\bibitem[{Naveed et~al.(2025)Naveed, Khan, Qiu, Saqib, Anwar, Usman, Akhtar, Barnes, and Mian}]{naveed2025comprehensive}
Humza Naveed, Asad~Ullah Khan, Shi Qiu, Muhammad Saqib, Saeed Anwar, Muhammad Usman, Naveed Akhtar, Nick Barnes, and Ajmal Mian. 2025.
\newblock A comprehensive overview of large language models.
\newblock \emph{ACM Transactions on Intelligent Systems and Technology}, 16(5):1--72.

\bibitem[{Nentidis et~al.(2024)Nentidis, Katsimpras, Krithara, Lima-L{\'o}pez, Farr{\'e}-Maduell, Krallinger, Loukachevitch, Davydova, Tutubalina, and Paliouras}]{nentidis2024overview}
Anastasios Nentidis, Georgios Katsimpras, Anastasia Krithara, Salvador Lima-L{\'o}pez, Eul{\`a}lia Farr{\'e}-Maduell, Martin Krallinger, Natalia Loukachevitch, Vera Davydova, Elena Tutubalina, and Georgios Paliouras. 2024.
\newblock Overview of bioasq 2024: the twelfth bioasq challenge on large-scale biomedical semantic indexing and question answering.
\newblock In \emph{International Conference of the Cross-Language Evaluation Forum for European Languages}, pages 3--27. Springer.

\bibitem[{OLMo et~al.(2025)OLMo, Walsh, Soldaini, Groeneveld, Lo, Arora, Bhagia, Gu, Huang, Jordan, Lambert, Schwenk, Tafjord, Anderson, Atkinson, Brahman, Clark, Dasigi, Dziri, Guerquin, Ivison, Koh, Liu, Malik, Merrill, Miranda, Morrison, Murray, Nam, Pyatkin, Rangapur, Schmitz, Skjonsberg, Wadden, Wilhelm, Wilson, Zettlemoyer, Farhadi, Smith, and Hajishirzi}]{olmo20252olmo2furious}
Team OLMo, Pete Walsh, Luca Soldaini, Dirk Groeneveld, Kyle Lo, Shane Arora, Akshita Bhagia, Yuling Gu, Shengyi Huang, Matt Jordan, Nathan Lambert, Dustin Schwenk, Oyvind Tafjord, Taira Anderson, David Atkinson, Faeze Brahman, Christopher Clark, Pradeep Dasigi, Nouha Dziri, Michal Guerquin, Hamish Ivison, Pang~Wei Koh, Jiacheng Liu, Saumya Malik, William Merrill, Lester James~V. Miranda, Jacob Morrison, Tyler Murray, Crystal Nam, Valentina Pyatkin, Aman Rangapur, Michael Schmitz, Sam Skjonsberg, David Wadden, Christopher Wilhelm, Michael Wilson, Luke Zettlemoyer, Ali Farhadi, Noah~A. Smith, and Hannaneh Hajishirzi. 2025.
\newblock \href {https://arxiv.org/abs/2501.00656} {2 olmo 2 furious}.
\newblock \emph{Preprint}, arXiv:2501.00656.

\bibitem[{Ong et~al.(2024)Ong, Chang, William, Butte, Shah, Chew, Liu, Doshi-Velez, Lu, Savulescu, and Ting}]{ong2024nejm}
Jasmine Chiat~Ling Ong, Shelley Yin-Hsi Chang, Wasswa William, Atul~J. Butte, Nigam~H. Shah, Lita Sui~Tjien Chew, Nan Liu, Finale Doshi-Velez, Wei Lu, Julian Savulescu, and Daniel Shu~Wei Ting. 2024.
\newblock \href {https://doi.org/10.1056/AIra2400038} {Medical ethics of large language models in medicine}.
\newblock \emph{NEJM AI}, 1(7):AIra2400038.

\bibitem[{Park et~al.(2025)Park, Yoon, Park, Lee, Jeong, and Kang}]{park2025chroknowledge}
Yein Park, Chanwoong Yoon, Jungwoo Park, Donghyeon Lee, Minbyul Jeong, and Jaewoo Kang. 2025.
\newblock \href {https://openreview.net/forum?id=whaO3482bs} {Chroknowledge: Unveiling chronological knowledge of language models in multiple domains}.
\newblock In \emph{The Thirteenth International Conference on Learning Representations}.

\bibitem[{Shi et~al.(2024)Shi, Xu, Wang, Qin, Wang, Wang, Wang, Ebrahimi, and Wang}]{shi2024continual}
Haizhou Shi, Zihao Xu, Hengyi Wang, Weiyi Qin, Wenyuan Wang, Yibin Wang, Zifeng Wang, Sayna Ebrahimi, and Hao Wang. 2024.
\newblock Continual learning of large language models: A comprehensive survey.
\newblock \emph{ACM Computing Surveys}.

\bibitem[{Singhal et~al.(2023)Singhal, Azizi, Tu, Mahdavi, Wei, Chung, Scales, Tanwani, Cole-Lewis, Pfohl et~al.}]{singhal2023large}
Karan Singhal, Shekoofeh Azizi, Tao Tu, S~Sara Mahdavi, Jason Wei, Hyung~Won Chung, Nathan Scales, Ajay Tanwani, Heather Cole-Lewis, Stephen Pfohl, et~al. 2023.
\newblock Large language models encode clinical knowledge.
\newblock \emph{Nature}, 620(7972):172--180.

\bibitem[{Singhal et~al.(2025)Singhal, Tu, Gottweis, Sayres, Wulczyn, Amin, Hou, Clark, Pfohl, Cole-Lewis et~al.}]{singhal2025toward}
Karan Singhal, Tao Tu, Juraj Gottweis, Rory Sayres, Ellery Wulczyn, Mohamed Amin, Le~Hou, Kevin Clark, Stephen~R Pfohl, Heather Cole-Lewis, et~al. 2025.
\newblock Toward expert-level medical question answering with large language models.
\newblock \emph{Nature Medicine}, pages 1--8.

\bibitem[{Soldaini et~al.(2024)Soldaini, Kinney, Bhagia, Schwenk, Atkinson, Authur, Bogin, Chandu, Dumas, Elazar, Hofmann, Jha, Kumar, Lucy, Lyu, Lambert, Magnusson, Morrison, Muennighoff, Naik, Nam, Peters, Ravichander, Richardson, Shen, Strubell, Subramani, Tafjord, Walsh, Zettlemoyer, Smith, Hajishirzi, Beltagy, Groeneveld, Dodge, and Lo}]{soldaini-etal-2024-dolma}
Luca Soldaini, Rodney Kinney, Akshita Bhagia, Dustin Schwenk, David Atkinson, Russell Authur, Ben Bogin, Khyathi Chandu, Jennifer Dumas, Yanai Elazar, Valentin Hofmann, Ananya Jha, Sachin Kumar, Li~Lucy, Xinxi Lyu, Nathan Lambert, Ian Magnusson, Jacob Morrison, Niklas Muennighoff, Aakanksha Naik, Crystal Nam, Matthew Peters, Abhilasha Ravichander, Kyle Richardson, Zejiang Shen, Emma Strubell, Nishant Subramani, Oyvind Tafjord, Evan Walsh, Luke Zettlemoyer, Noah Smith, Hannaneh Hajishirzi, Iz~Beltagy, Dirk Groeneveld, Jesse Dodge, and Kyle Lo. 2024.
\newblock \href {https://doi.org/10.18653/v1/2024.acl-long.840} {Dolma: an open corpus of three trillion tokens for language model pretraining research}.
\newblock In \emph{Proceedings of the 62nd Annual Meeting of the Association for Computational Linguistics (Volume 1: Long Papers)}, pages 15725--15788, Bangkok, Thailand. Association for Computational Linguistics.

\bibitem[{Soldaini and Lo(2023)}]{peS2o}
Luca Soldaini and Kyle Lo. 2023.
\newblock {peS2o (Pretraining Efficiently on S2ORC) Dataset}.
\newblock Technical report, {Allen Institute for AI}.
\newblock ODC-By, \url{https://github.com/allenai/pes2o}.

\bibitem[{Subramanian et~al.(2024)Subramanian, Schlegel, Ramesh~Kashyap, Nguyen, Dwivedi, and Winkler}]{subramanian-etal-2024-qalm}
Anand Subramanian, Viktor Schlegel, Abhinav Ramesh~Kashyap, Thanh-Tung Nguyen, Vijay~Prakash Dwivedi, and Stefan Winkler. 2024.
\newblock \href {https://doi.org/10.18653/v1/2024.findings-acl.238} {{M}-{QALM}: A benchmark to assess clinical reading comprehension and knowledge recall in large language models via question answering}.
\newblock In \emph{Findings of the Association for Computational Linguistics: ACL 2024}, pages 4002--4042, Bangkok, Thailand. Association for Computational Linguistics.

\bibitem[{Thirunavukarasu et~al.(2023)Thirunavukarasu, Ting, Elangovan, Gutierrez, Tan, and Ting}]{thirunavukarasu2023large}
Arun~James Thirunavukarasu, Darren Shu~Jeng Ting, Kabilan Elangovan, Laura Gutierrez, Ting~Fang Tan, and Daniel Shu~Wei Ting. 2023.
\newblock Large language models in medicine.
\newblock \emph{Nature medicine}, 29(8):1930--1940.

\bibitem[{Tirumala et~al.(2022)Tirumala, Markosyan, Zettlemoyer, and Aghajanyan}]{tirumala2022memorization}
Kushal Tirumala, Aram~H. Markosyan, Luke Zettlemoyer, and Armen Aghajanyan. 2022.
\newblock \href {https://openreview.net/forum?id=u3vEuRr08MT} {Memorization without overfitting: Analyzing the training dynamics of large language models}.
\newblock In \emph{Advances in Neural Information Processing Systems}.

\bibitem[{Vladika et~al.(2024{\natexlab{a}})Vladika, Schneider, and Matthes}]{vladika-etal-2024-healthfc}
Juraj Vladika, Phillip Schneider, and Florian Matthes. 2024{\natexlab{a}}.
\newblock \href {https://aclanthology.org/2024.lrec-main.709/} {{H}ealth{FC}: Verifying health claims with evidence-based medical fact-checking}.
\newblock In \emph{Proceedings of the 2024 Joint International Conference on Computational Linguistics, Language Resources and Evaluation (LREC-COLING 2024)}, pages 8095--8107, Torino, Italia. ELRA and ICCL.

\bibitem[{Vladika et~al.(2024{\natexlab{b}})Vladika, Schneider, and Matthes}]{vladika-etal-2024-medreqal}
Juraj Vladika, Phillip Schneider, and Florian Matthes. 2024{\natexlab{b}}.
\newblock \href {https://doi.org/10.18653/v1/2024.findings-acl.860} {{M}ed{REQAL}: Examining medical knowledge recall of large language models via question answering}.
\newblock In \emph{Findings of the Association for Computational Linguistics: ACL 2024}, pages 14459--14469, Bangkok, Thailand. Association for Computational Linguistics.

\bibitem[{Vu et~al.(2024)Vu, Iyyer, Wang, Constant, Wei, Wei, Tar, Sung, Zhou, Le, and Luong}]{vu-etal-2024-freshllms}
Tu~Vu, Mohit Iyyer, Xuezhi Wang, Noah Constant, Jerry Wei, Jason Wei, Chris Tar, Yun-Hsuan Sung, Denny Zhou, Quoc Le, and Thang Luong. 2024.
\newblock \href {https://doi.org/10.18653/v1/2024.findings-acl.813} {{F}resh{LLM}s: Refreshing large language models with search engine augmentation}.
\newblock In \emph{Findings of the Association for Computational Linguistics: ACL 2024}, pages 13697--13720, Bangkok, Thailand. Association for Computational Linguistics.

\bibitem[{Wallace et~al.(2022)Wallace, Barak, Truong, and Parker}]{wallace2022hierarchy}
Sowdhamini~S Wallace, Gal Barak, Grace Truong, and Michelle~W Parker. 2022.
\newblock Hierarchy of evidence within the medical literature.
\newblock \emph{Hospital Pediatrics}, 12(8):745--750.

\bibitem[{Wang et~al.(2023)Wang, Xie, Pei, Chen, Tiwari, Li, and Fu}]{wang2023pre}
Benyou Wang, Qianqian Xie, Jiahuan Pei, Zhihong Chen, Prayag Tiwari, Zhao Li, and Jie Fu. 2023.
\newblock Pre-trained language models in biomedical domain: A systematic survey.
\newblock \emph{ACM Computing Surveys}, 56(3):1--52.

\bibitem[{Wang et~al.(2024{\natexlab{a}})Wang, Zhu, Liu, Zheng, Chen, and Li}]{wang2024knowledge}
Song Wang, Yaochen Zhu, Haochen Liu, Zaiyi Zheng, Chen Chen, and Jundong Li. 2024{\natexlab{a}}.
\newblock Knowledge editing for large language models: A survey.
\newblock \emph{ACM Computing Surveys}, 57(3):1--37.

\bibitem[{Wang et~al.(2024{\natexlab{b}})Wang, Feng, Wang, Shi, Balachandran, He, and Tsvetkov}]{wang2024resolving}
Yike Wang, Shangbin Feng, Heng Wang, Weijia Shi, Vidhisha Balachandran, Tianxing He, and Yulia Tsvetkov. 2024{\natexlab{b}}.
\newblock \href {https://openreview.net/forum?id=ptvV5HGTNN} {Resolving knowledge conflicts in large language models}.
\newblock In \emph{First Conference on Language Modeling}.

\bibitem[{White(2020)}]{white2020pubmed}
Jacob White. 2020.
\newblock Pubmed 2.0.
\newblock \emph{Medical reference services quarterly}, 39(4):382--387.

\bibitem[{Wu et~al.(2023)Wu, Lin, Zhang, Zhang, Wang, and Xie}]{wu2023pmcllama}
Chaoyi Wu, Weixiong Lin, Xiaoman Zhang, Ya~Zhang, Yanfeng Wang, and Weidi Xie. 2023.
\newblock \href {https://arxiv.org/abs/2304.14454} {Pmc-llama: Towards building open-source language models for medicine}.
\newblock \emph{Preprint}, arXiv:2304.14454.

\bibitem[{W{\"u}hrl et~al.(2024)W{\"u}hrl, Wright, Klinger, and Augenstein}]{wuhrl2024understanding}
Amelie W{\"u}hrl, Dustin Wright, Roman Klinger, and Isabelle Augenstein. 2024.
\newblock Understanding fine-grained distortions in reports of scientific findings.
\newblock In \emph{Findings of the Association for Computational Linguistics ACL 2024}, pages 6175--6191.

\bibitem[{Xu et~al.(2024)Xu, Qi, Guo, Wang, Wang, Zhang, and Xu}]{xu-etal-2024-knowledge-conflicts}
Rongwu Xu, Zehan Qi, Zhijiang Guo, Cunxiang Wang, Hongru Wang, Yue Zhang, and Wei Xu. 2024.
\newblock \href {https://doi.org/10.18653/v1/2024.emnlp-main.486} {Knowledge conflicts for {LLM}s: A survey}.
\newblock In \emph{Proceedings of the 2024 Conference on Empirical Methods in Natural Language Processing}, pages 8541--8565, Miami, Florida, USA. Association for Computational Linguistics.

\bibitem[{Yang et~al.(2024)Yang, Wen, Qu, Chen, Xiang, Chen, and Yao}]{yang2024memorization}
Xinyu Yang, Zichen Wen, Wenjie Qu, Zhaorun Chen, Zhiying Xiang, Beidi Chen, and Huaxiu Yao. 2024.
\newblock \href {https://openreview.net/forum?id=KmW8WkCKRx} {Memorization and privacy risks in domain-specific large language models}.
\newblock In \emph{ICLR 2024 Workshop on Reliable and Responsible Foundation Models}.

\bibitem[{Yao et~al.(2024)Yao, Xu, and YangLiu}]{NEURIPS2024_be52acf6}
Yuanshun Yao, Xiaojun Xu, and YangLiu. 2024.
\newblock \href {https://proceedings.neurips.cc/paper_files/paper/2024/file/be52acf6bccf4a8c0a90fe2f5cfcead3-Paper-Conference.pdf} {Large language model unlearning}.
\newblock In \emph{Advances in Neural Information Processing Systems}, volume~37, pages 105425--105475. Curran Associates, Inc.

\bibitem[{Yu et~al.(2024)Yu, Ping, Liu, Wang, You, Zhang, Shoeybi, and Catanzaro}]{yu2024rankrag}
Yue Yu, Wei Ping, Zihan Liu, Boxin Wang, Jiaxuan You, Chao Zhang, Mohammad Shoeybi, and Bryan Catanzaro. 2024.
\newblock \href {https://openreview.net/forum?id=S1fc92uemC} {Rank{RAG}: Unifying context ranking with retrieval-augmented generation in {LLM}s}.
\newblock In \emph{The Thirty-eighth Annual Conference on Neural Information Processing Systems}.

\bibitem[{Zhang et~al.(2023)Zhang, Fang, Chen, Namazi-Rad, and Wang}]{zhang-etal-2023-large}
Zihan Zhang, Meng Fang, Ling Chen, Mohammad-Reza Namazi-Rad, and Jun Wang. 2023.
\newblock \href {https://doi.org/10.18653/v1/2023.emnlp-main.516} {How do large language models capture the ever-changing world knowledge? a review of recent advances}.
\newblock In \emph{Proceedings of the 2023 Conference on Empirical Methods in Natural Language Processing}, pages 8289--8311, Singapore. Association for Computational Linguistics.

\bibitem[{Zhang et~al.(2025)Zhang, Lin, Zheng, and Wu}]{zhang2025much}
Ziheng Zhang, Zhenxi Lin, Yefeng Zheng, and Xian Wu. 2025.
\newblock How much medical knowledge do llms have? an evaluation of medical knowledge coverage for llms.
\newblock In \emph{Proceedings of the ACM on Web Conference 2025}, pages 5330--5341.

\end{thebibliography}

\appendix

\section{Simple RAG Improvement}
\label{sec:appendix_rag}

Common ways to address the outdated knowledge with \textit{knowledge editing} include continual learning methods and external search augmentation. We do a simple experiment using a retrieval-augmented method. For each of the 512 questions in MedChangeQA, we query the PubMed API and take the abstract of the Top 1 result, and append it to the main prompt as an additional context. Results are shown in Table~\ref{tab:RAG}. This improves the F1 scores by a margin of 3--16 and partially closes the gap, but still leaves a lot of room for improvement. This serves as a simple demonstration of one way to address the outdated knowledge -- future work could focus on retrieving more documents, using structured and focused search queries (like searching for SLRs), semantic search, graph RAG, learning to re-rank and avoid conflicts, etc. Additionally, methods of continual learning and fine-tuning could be used, with MedChangeQA serving as a testbed to measure the rate of success of the proposed techniques.

\begin{table}[htpb]
\centering
\small
\begin{tabular}{l|ccc|c}
\hline
           & \textbf{P}  & \textbf{R} & \textbf{F1} & \textbf{Improv.~F1} \\ \hline
\textbf{GPT 4o} & 43.4  &  40.2 & 39.8 & +8.7 \\ 
\textbf{Mistral} & 47.5 & 41.5 & 39.6   & +5.9 \\ 
\textbf{Llama 3.3}   &   44.1   &   42.3  &  38.8   & +4.7  \\
\textbf{Qwen} &  43.3 &  43.7  &   42.2 &  +16.2 \\ 
\textbf{Deepseek}   &   40.7  &  39.3    &   35.4   & +3.2 \\ \hline
\end{tabular}


\caption{\label{tab:RAG} Performance improvements with the abstract of the top PubMed paper included in prompt.}
\end{table}

\section{Pre-Training Data of LLMs}
\label{sec:training_data}

This section provides an overview of what is publicly known about pre-training data for the used LLMs, as reported in their technical reports or official documentation:
\begin{itemize}
    \item \textbf{Llama 3.3}: Pretrained on approximately 15 trillion tokens of data sourced from publicly available online sources. The exact composition and breakdown of the dataset are not detailed, but Meta emphasizes that the data is "a new mix" of public internet data. The data cutoff for pretraining is December 2023.
    \item \textbf{Mistral 24B} The official technical report and available documentation do not provide explicit details about the pre-training corpus for Mistral 24B. However, Mistral's models are generally known to be trained on large-scale, diverse datasets, often including filtered web data, code, and other standard sources, but no specifics are publicly disclosed for the 24B version in the sources provided.
    \item \textbf{GPT 4o}  It was trained on data up to October 2023, sourced from a "wide variety of materials," including: (a) publicly available data (web pages, ML datasets, common crawls), (b) proprietary data (obtained via data partnerships, e.g., paywalled content, archives, metadata), (c) key dataset components (web data, code and math data, multimodal data). The dataset underwent safety filtering to remove harmful content, personal information, and explicit material. OpenAI does not provide a detailed breakdown of dataset proportions or specific sources.
    \item \textbf{Qwen 2.5} It was trained on up to 18 trillion tokens of data. The dataset is described as "large-scale" and "high-quality," but the technical report does not specify the exact sources. The data is designed to provide a strong foundation for common sense, expert knowledge, and reasoning. Qwen 2.5 also supports multilingual capabilities across more than 29 languages.
    \item \textbf{DeepSeek V3} It was trained on 14.8 trillion tokens of "diverse and high-quality" data. The dataset construction focused on: an increased ratio of mathematical and programming samples, multilingual coverage, and a data processing pipeline optimized for diversity and minimal redundancy. The technical report does not provide a granular breakdown of data sources but highlights the focus on math, code, and multilingual content.
    \item \textbf{OLMo 2} is trained on the Dolma corpus \cite{soldaini-etal-2024-dolma}, a fully open dataset containing around 3 trillion tokens. This is a high-level breakdown of the composition of the pre-training corpus:
    Common Crawl (2,479 billion tokens), 
    GitHub (411 billion tokens),
    Reddit (89 billion tokens), 
    Semantic Scholar (70 billion tokens), 
    Project Gutenberg (6.0 billion tokens), 
    Wikipedia and Wikibooks (4.3 billion tokens). 
    \item \textbf{PMC-Llama} and \textbf{BioMistral} use the base models of Llama and Mistral as described before, but were then further pre-trained on abstracts of biomedical publications from PubMed and full publications from PubMed Central. As described in our paper, PubMed contains all the abstracts of Cochrane systematic reviews, which means all the Cochrane SLRs from our dataset were a part of training.
\end{itemize}

\begin{table*}[thb]
\centering
\small

\resizebox{\textwidth}{!}{

\begin{tabular}{p{22mm}|rrrrrrrr}
\hline
& \textbf{Llama 3.3} & \textbf{Mistral} & \textbf{GPT-4o} &\textbf{ Qwen 2.5} & \textbf{Deepseek} & \textbf{OLMo 2} & \textbf{BioMistral} & \textbf{PMC-L} \\
\hline
"\textbf{Cochrane}" & 51 & 783 & 2 & 629 & 901 & 283 & 2067 & 2344 \\
\hline
 "\textbf{systematic review}" & 221 & 1664 & 623 & 3194 & 3046 & 531 & 3990 & 4956 \\
\hline
"\textbf{meta-analysis}" & 844 & 3511 & 714 & 4180 & 2776 & 981 & 4253 & 4618 \\
\hline
"\textbf{journal}" & 53 & 689 & 7 & 4620 & 448 & 196 & 574 & 624 \\
\hline
"\textbf{studies}" & 7024 & 12419 & 13493 & 12516 & 13421 & 6615 & 7598 & 9720 \\
\hline
\end{tabular}

}

\caption{Number of answers (out of 16,501) from each tested LLM where the respective terms were mentioned. This shows the tendency to refer to and cite relevant medical studies that were memorized during pre-training. Two biomedical models, which were pre-trained on biomedical publications, also refer to specific studies the most.}
\label{tab:mentions}

\end{table*}

\newpage

\section{Mentions of Common Terms}
Table \ref{tab:mentions} show the count of how many times were some common terms mentioned in answers of the eight evaluated LLMs to 16,501 questions from the MedRevQA dataset. These terms include those that signal a mention of specific studies used as a reference for the answer, such as \textit{systematic review}, \textit{meta-analysis}, \textit{journal}, or \textit{Cochrane} since that is the publishing organization, from which our questions and labels originate. We also included the generic term \textit{studies}, which is often mentioned in those answers that do not refer to specific studies but only give a general statement such as "\textit{Many studies have shown that...}". The use of this generic term was especially common in GPT-4o, which mentioned specific studies the least. The two biomedical LLMs, BioMistral and PMC-Llama, which were further pre-trained on full texts of biomedical publications from PubMed Central, also tended to cite specific studies the most.

\section{Prompts and Examples}
\label{sec:appendix}
\begin{figure*}[h]
    \centering
    \includegraphics[width=0.99\linewidth]{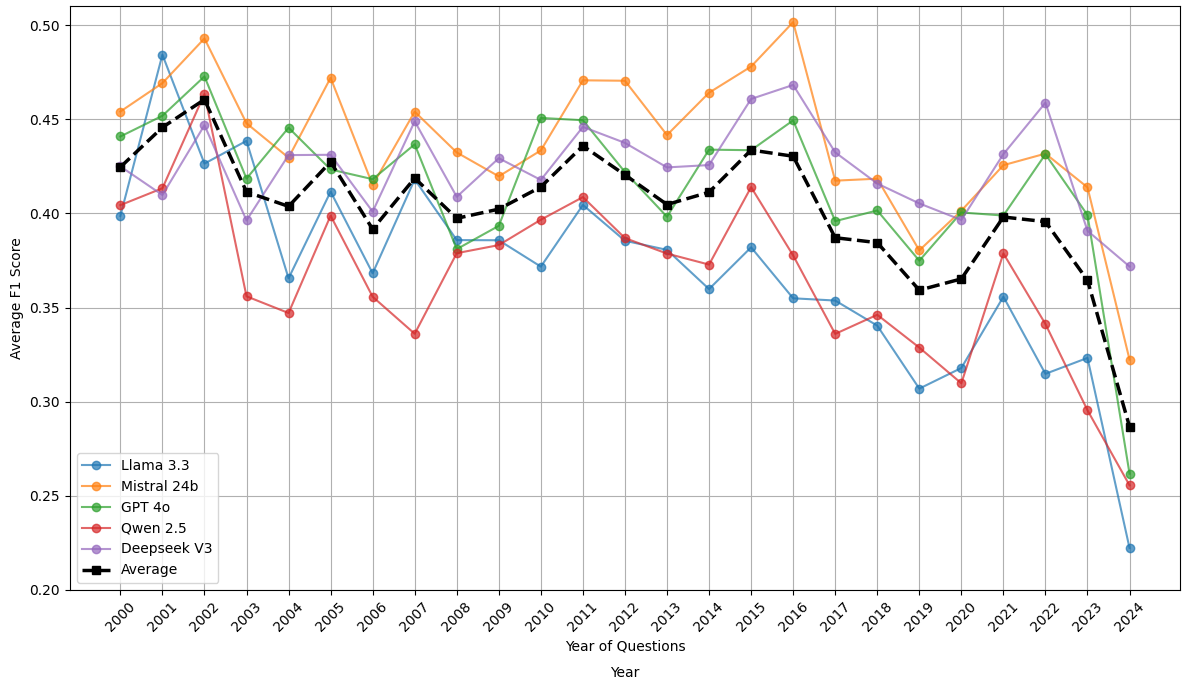}
    \caption{Larger version of the previous figure (Fig. \ref{fig:f1_years}): Average F1-Macro performance for questions originating from each year in the dataset across five LLMs, showing decline in more recent years.}
    \label{fig:full_f1_years}
\end{figure*}

This Appendix section provides additional material for the study, including the model prompts in full length (Tables~\ref{tab:data_prompts} and \ref{tab:llm_prompts}) and example questions and model answers (Tables~\ref{tab:outdated} and \ref{tab:citewrong}). 
Figure \ref{fig:full_f1_years} shows a larger version of the plot of the average F1 score for tested LLMs on questions over the years.

\begin{table*}[htpb]
\centering
\begin{tabular}{p{28mm}p{122mm}}
\hline
\textbf{Use Case} & \textbf{Prompt Content}\\
Question \& Label generation & \verb|SYSTEM:| You're a helpful assistant. Your task is to help with generating questions and labels in the medical and clinical domain.
\newline \verb|USER| You will be given an excerpt of an abstract of a clinical systematic review. Based on the given background, objectives, and author's conclusions, generate only ONE SINGLE question, answerable with yes/no/uncertain, that sums up the main medical objective that was investigated. Please keep the question short and general and use the "Objectives" section to construct the question. The question should be about a general medical hypothesis, not about this specific review.
\newline
    Afterwards, please also give a label for the author's conclusions. The label tries to answer the objective by looking at the conclusion. The label may be ONLY from one of the following three: (1) SUPPORTED; (2) REFUTED; (3) NOT ENOUGH INFORMATION. Do not try to make up a new label. Please only select the third label if not enough evidence was found to reach a verdict, not if the certainty of the conclusion is low! Please aim to predict "SUPPORTED" or "REFUTED" even if the certainty of these conclusions by the authors is low or moderate.
\newline
    Please structure the output in two lines, as:
\newline    
    QUESTION: (question)
\newline
    LABEL: (label)
\newline
    The documents begins now.

    \\ \hline

\end{tabular}
\caption{Overview of applied prompts for data generation and annotation.}
\label{tab:data_prompts}
\end{table*}

\begin{table*}[htpb]
\centering
\begin{tabular}{p{29mm}p{120mm}}
\hline
          \textbf{Model} & \textbf{Prompt}  \\ \hline

          \textbf{PMC-LLaMa} &  Below is an instruction that describes a task, paired with an input that provides further context. Write a response that appropriately completes the request.
\newline
\#\#\# Instruction: Based on your best current knowledge, please answer the following medical question. If you think there is not enough evidence to answer, then say so. Please answer the question with "SUPPORTED" or "REFUTED" or "NOT ENOUGH INFORMATION". Briefly explain your answer.
\newline
\#\#\# Input: \{question\}

\#\#\# Response:

  \\ \hline

\textbf{BioMistral} & <s>[INST]  Based on your knowledge, please answer this clinical question only with SUPPORTED (if the question is supported by the clinical research) or REFUTED (if the hypothesis is refuted by the current clinical research) or NOT ENOUGH INFORMATION (if there is insufficient evidence for the question in current research). Please give your output in form of LABEL: (label) . Briefly explain your answer.
\newline
QUESTION: \{question\}
\newline
[/INST]

\\ \hline
\textbf{Mistral 24B, Llama 3.3, GPT-4o, Qwen 2.5, DeepSeek-V3} & \verb|SYSTEM| You are an AI assistant helping answer clinical and medical questions based on your best knowledge.
\newline
\verb|AGENT| Please answer this clinical question only with SUPPORTED (if the question is supported by the clinical research) or REFUTED (if the hypothesis is refuted by the current clinical research) or NOT ENOUGH INFORMATION (if there is insufficient evidence for the question in current research). Please give your output in form of LABEL: (label) . Briefly explain your answer.
\newline
QUESTION: \{question\}
\\  \hline
\end{tabular}
%

\caption{\label{tab:llm_prompts} Prompts used for each of the LLMs in experiments.}
\end{table*}

\begin{table*}[h!]
\centering

\begin{tabular}{p{19mm}p{132mm}}
\hline
\textbf{Question} &  \small Is routine malaria chemoprophylaxis beneficial for people with sickle cell disease in malaria-endemic areas?  \\[0.5ex]

\textbf{Conclusion (Latest)}  & \small  (...) It is beneficial to give routine malaria chemoprophylaxis in sickle cell disease in areas where malaria is endemic. (Oniyangi, 2019) [\textbf{label}: \textcolor{teal}{\textbf{Supported}}]  \\[0.5ex]

\textbf{Conclusion (Outdated)}  & \small  (...)  There is very little direct evidence to support or refute giving routine chemoprophylaxis in sickle cell disease in areas where malaria is endemic. (Oniyangi, 2006) [\textbf{label}: \textcolor{blue}{\textbf{Not Enough Information}}]  \\[0.5ex]

\hline
\textbf{GPT-4o} & 
\small Routine malaria chemoprophylaxis is beneficial for people with sickle cell disease in malaria-endemic areas. Clinical research indicates that individuals with sickle cell disease are at increased risk for severe malaria and its complications. Chemoprophylaxis can reduce the incidence of malaria and its associated morbidity in this vulnerable population, thereby improving health outcomes. 
 [\textbf{predicted}: \textcolor{teal}{\textbf{Supported}}]  \\[0.5ex]

\textbf{Llama 3.3} & 
\small Current clinical research suggests that routine malaria chemoprophylaxis is not recommended for people with sickle cell disease in malaria-endemic areas, as the risks associated with long-term prophylaxis may outweigh the benefits, and other preventive measures such as insecticide-treated bed nets and prompt treatment of malaria episodes are preferred. [\textbf{predicted}: \textcolor{red}{\textbf{Refuted}}] \\[0.5ex]

 \textbf{Mistral 24B} & 
\small Routine malaria chemoprophylaxis is generally recommended for people with sickle cell disease living in malaria-endemic areas. This is supported by clinical research indicating that malaria infections can trigger sickle cell crises and other complications. Chemoprophylaxis helps reduce the risk of malaria, thereby decreasing the frequency of these crises and improving overall health outcomes.  [\textbf{predicted}: \textcolor{teal}{\textbf{Supported}}] 

    \\ \hline

\end{tabular}
\caption{Example of a question and two labels, one outdated and one recent from our dataset. Answers show how GPT and Mistral capture the latest and up-to-date knowledge for this question, while Llama predicted incorrect and outdated recommendations.}
\label{tab:outdated}
\end{table*}


\begin{table*}[h]
\centering

\begin{tabular}{p{19mm}p{132mm}}
\hline
\textbf{Question} &  \small Do probiotics reduce the risk of necrotising enterocolitis in very preterm or low birth weight infants?  \\[0.5ex]

\textbf{Conclusion (Outdated)}  & \small  Enteral supplementation of probiotics prevents severe NEC and all cause mortality in preterm infants. Our updated review of available evidence strongly supports a change in practice. (...) (AlFaleh et al., 2014) [\textbf{label}: \textcolor{teal}{\textbf{Supported}}]  \\[0.5ex]

\textbf{Conclusion (Latest)}  & \small Given the low to moderate certainty of evidence for the effects of probiotic supplements on the risk of NEC and associated morbidity and mortality for very preterm or VLBW infants, and particularly for extremely preterm or ELBW infants, there is a need for further large, high-quality trials to provide evidence of sufficient validity and applicability to inform policy and practice. (Sharif et al., 2023) [\textbf{label}: \textcolor{blue}{\textbf{Not Enough Information}}]  \\[0.5ex]

\hline

 \textbf{Mistral 24B} & 
\small Several clinical studies and meta-analyses have shown that probiotic supplementation in very preterm or low birth weight infants can reduce the risk of necrotizing enterocolitis (NEC). A \textbf{Cochrane review from 2014} has supported this finding, indicating that probiotics can be beneficial in preventing NEC in this vulnerable population.  [\textbf{predicted}: \textcolor{teal}{\textbf{Supported}}] 

    \\ \hline

\end{tabular}
\caption{Example of a study objective that had the \textit{Supported} label in its 2014 version and an updated \textit{Not enough information} label in its 2023 version. Mistral cited the more than decade-old study to supports its answer.}
\label{tab:citewrong}
\end{table*}


\end{document}